\documentclass{article}
\usepackage{fancyhdr}
\usepackage{spconf,amsmath,graphicx}
\usepackage{amsmath,amssymb,graphicx}
\usepackage{amsfonts}
\usepackage{url}
\usepackage{algpseudocode}
\usepackage{booktabs}
\usepackage{adjustbox}
\usepackage{threeparttable}
\usepackage{multirow}
\usepackage[table]{xcolor}%
\usepackage{subcaption}
\usepackage{array}
\usepackage{amsthm,amsmath,amssymb}
\usepackage{mathrsfs}
\usepackage{arydshln} 
\usepackage{makecell}
\usepackage{psfrag}
\usepackage{epsfig}
\usepackage{rotating}
\usepackage{colortbl}

\renewcommand{\thetable}{\arabic{table}}

\usepackage{hyperref}
\hypersetup{hypertex=true,
            colorlinks=true,
            linkcolor=red,
            anchorcolor=red,
            citecolor=green}

\fancypagestyle{firstpage}{
  \fancyhf{}
}
\title{Controllable-Continuous Color Editing in Diffusion Model via Color Mapping}
\name{Yuqi Yang$^1$, Dongliang Chang$^{1,*}$, Yuanchen Fang$^1$, Yi-Zhe SonG$^2$, Zhanyu Ma$^1$, and Jun Guo$^1$
\thanks{*: Corresponding author: changdongliang@pris-cv.cn}}
\address{$^{1}$School of Artificial Intelligence, Beijing University of Posts and Telecommunications \\$^{2}$SketchX, CVSSP, University of Surrey
}
\begin{document}
%\ninept
%
\maketitle
\thispagestyle{firstpage}

\begin{abstract}
In recent years, text-driven image editing has made significant progress. However, due to the inherent ambiguity and discreteness of natural language, color editing still faces challenges such as insufficient precision and difficulty in achieving continuous control. Although linearly interpolating the embedding vectors of different textual descriptions can guide the model to generate a sequence of images with varying colors, this approach lacks precise control over the range of color changes in the output images. Moreover, the relationship between the interpolation coefficient and the resulting image color is unknown and uncontrollable. To address these issues, we introduce a color mapping module that explicitly models the correspondence between the text embedding space and image RGB values. This module predicts the corresponding embedding vector based on a given RGB value, enabling precise color control of the generated images while maintaining semantic consistency. Users can specify a target RGB range to generate images with continuous color variations within the desired range, thereby achieving finer-grained, continuous, and controllable color editing. Experimental results demonstrate that our method performs well in terms of color continuity and controllability.

\end{abstract}
\begin{keywords}
Image Editing, Color Control, Continuous Editing, Diffusion Model
\end{keywords}

\vspace{-10pt}
\section{Introduction}
\label{sec:intro}
In recent years, with the rapid development of text-driven image generation technologies, users can edit image content through natural language descriptions~\cite{du2024demofusion, lin2024text, brack2024ledits++, chang2023making, tong2025reserve}. User demands have evolved toward fine-grained attribute editing, among which color attribute editing has garnered widespread attention due to its intuitive visual perception~\cite{yin2024benchmark, cheng2024zest, garifullin2025materialfusion, chang2023erudite}.

\begin{figure}[t]
\begin{center}
  \includegraphics[width=0.85\linewidth]{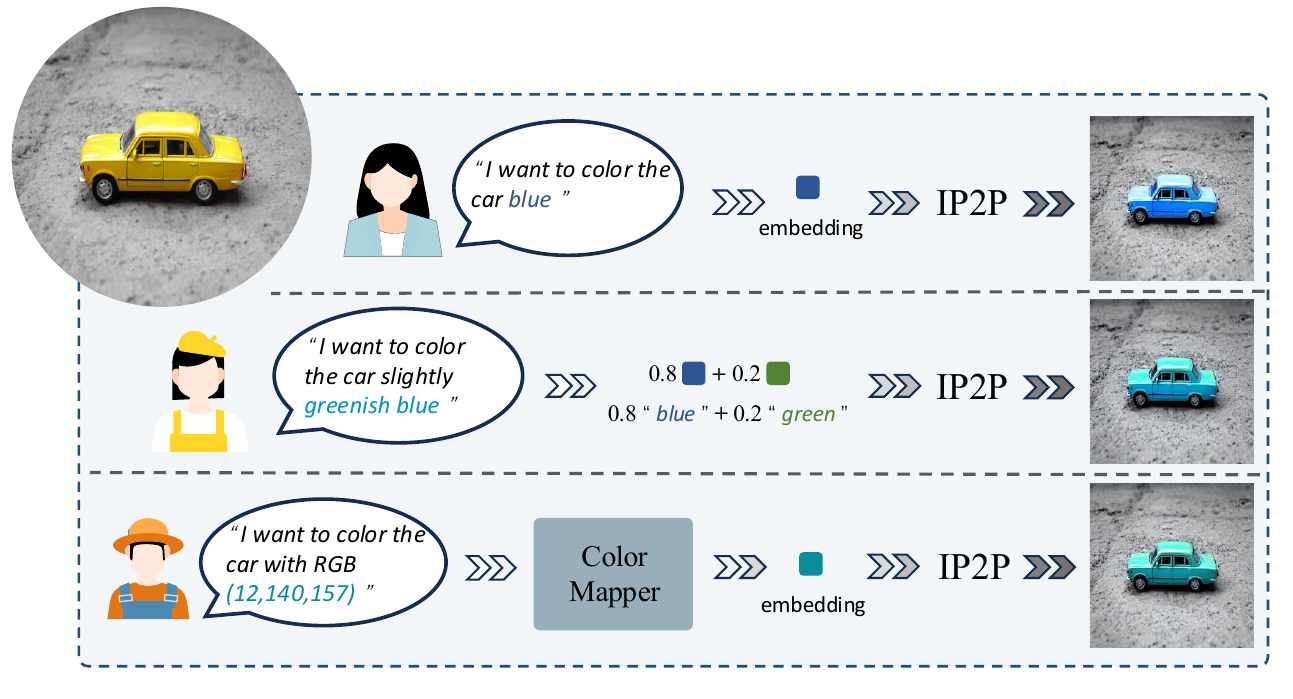}
\end{center}
    \vspace{-15pt}
  \caption{Different editing frameworks, the first row is vanilla IP2P, the second row is IP2P+Interpolation, the third row is Our method.}
  \label{fig1:insight}
  \vspace{-15pt}
\end{figure}

Prompt-to-Prompt~\cite{hertz2022prompt} achieves fine-grained editing of local regions in generated images by manipulating the attention maps of textual prompts during the diffusion process. InstructPix2Pix~\cite{brooks2023instructpix2pix} enables the modification of the target image by accepting natural language editing instructions, while Imagic~\cite{kawar2023imagic} fine-tunes a personalized diffusion model conditioned on a specific image and text, achieving high-fidelity image edits consistent with the target description. These approaches largely rely on natural language to express editing intentions. However, color is inherently a fine-grained and continuous attribute, and natural language has certain limitations in expressing color precisely, thus hindering fine-grained control in color editing.

\begin{figure*}[t]
\begin{center}
  \includegraphics[width=0.85\linewidth]{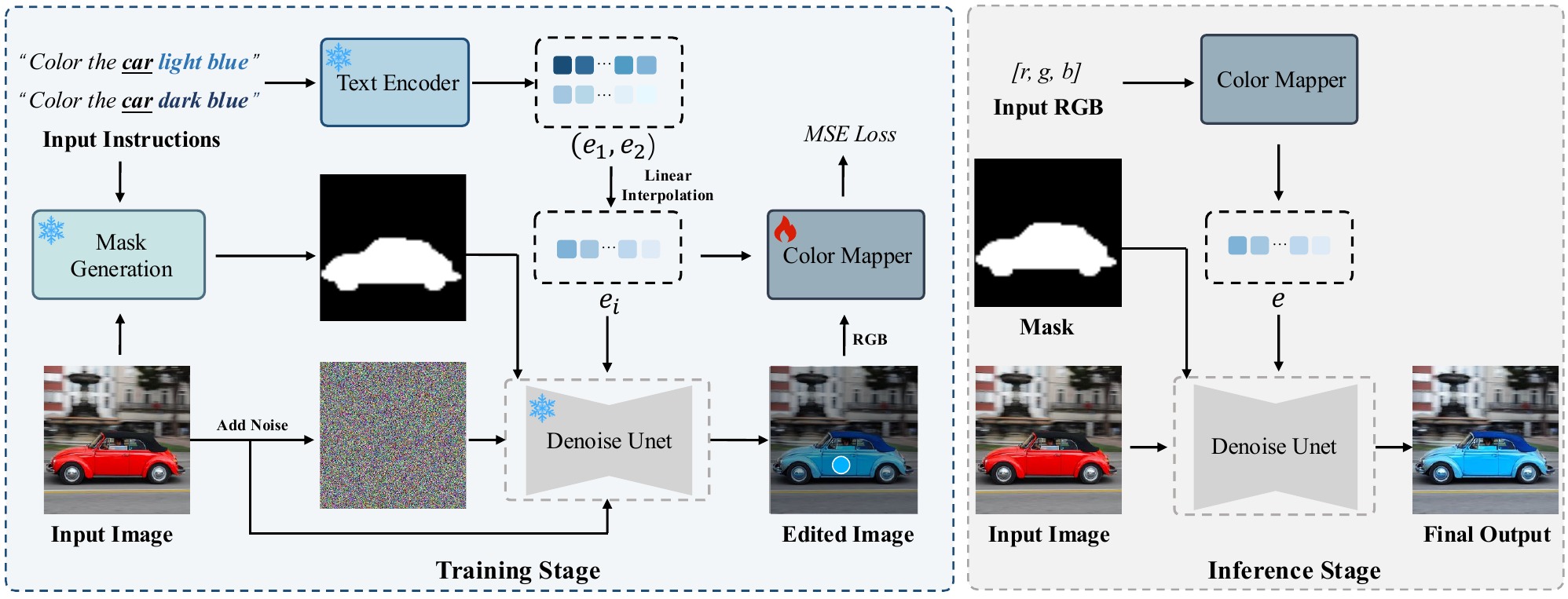}
\end{center}
\vspace{-10pt}
  \caption{{\bf Framework of the proposed method.} We first obtain the target editing region using Mask Generation to generate a binary mask. Given two color editing prompts representing different target colors, we perform linear interpolation between their embeddings to obtain intermediate color state embeddings, which are then used to guide the image generation. Next, we extract the RGB values of annotated pixels from the generated images and use them to train the Color Mapper, learning the mapping between RGB values and embeddings. During inference, users can input a desired RGB value to directly obtain the corresponding edited image.}
  \label{fig2:method}
  \vspace{-10pt}
\end{figure*}

On the one hand, natural language cannot describe colors as precisely as RGB values. For example, color descriptions such as ``red'', ``blue'', or ``yellow'' are often general and lack the specificity needed for precise editing. On the other hand, natural language is discrete and struggles to convey subtle changes in color within a continuous space. Phrases such as ``slightly greenish blue'' tend to be subjective and ambiguous, making them inadequate to guide precise color transformations. Methods such as~\cite{ye2023ip, yang2025adaptive, liu2024referring, du2023semi} have attempted to enhance the controllability of diffusion-based editing by introducing additional conditioning. However, these methods often rely on extra supervision or are not specifically designed for color control, and thus still fall short in meeting the demands of precise color editing.

Some studies~\cite{baumann2024continuous, radford2021learning, bian2024class, chen2022cross} have shown that diffusion models can interpret modified prompt embeddings—even those not corresponding to any specific word. Based on this insight, we first attempted to achieve accurate color control and smooth transitions by interpolating between embedding vectors of color-related text descriptions, as shown in Fig.~\ref{fig1:insight}. While this interpolation method can generate a sequence of images with gradual color changes by adjusting the interpolation coefficient, it lacks a clear and accurate mapping between the coefficient and the resulting RGB values. Consequently, we must blindly and repeatedly tweak the interpolation coefficient to alter the color range of the output images.

To address this limitation, we propose a color mapping-based method for controllable and continuous color editing. We introduce a color mapping module that models the relationship between the text embedding space and RGB values. By learning the correspondence between RGB values and textual embeddings during image generation, the module {\bf Color Mapper} can take user-specified target RGB values or ranges as input, compute the corresponding embedding vector, and guide the diffusion model to generate images with smoothly varying colors within the desired RGB range. This enables fine-grained, continuous, and controllable color editing. Moreover, to prevent unwanted changes in irrelevant regions of the image, we incorporate a mask to constrain the editing area, ensuring the preservation of the original image structure.

\vspace{-10pt}
\section{Method}
\label{sec:format}
In this section, we first introduce the preliminaries, then we leverage the prompt linear interpolation and the color mapper module. Finally, we introduce the inference process. Framework of the proposed method is shown in Fig.~\ref{fig2:method}.

\vspace{-10pt}
\subsection{Preliminaries}
{\bf InstrucutPix2Pix~\cite{brooks2023instructpix2pix}} The InstrucutPix2Pix (IP2P) method is built upon Stable Diffusion~\cite{rombach2022high} and leverages a pre-trained Variational Autoencoder~\cite{kingma2013auto} (VAE) to enhance the efficiency and generation quality of the diffusion model. The VAE consists of an encoder and a decoder. Each sample in IP2P dataset contains the original image $I$, the editing instruction $T$ and the corresponding edited image $I_e$, and supervised training is conducted on this dataset. During the training phase, given an edited image $I_e$, it is first encoded into a latent representation $z = \mathcal{E}(I_e)$, then, the noise sampled from a standard normal distribution $N(0,1)$ is added to form the latent variable $z_t$. The network $\epsilon_{\theta}$ is trained by minimizing the following diffusion loss function:
\begin{equation}
\begin{split}
  & L = \mathbb{E}_{\mathcal{E}(x), \mathcal{E}(c_I), c_T, \epsilon \sim \mathcal{N}(0,1), t} 
\left[ \left\| \epsilon - \epsilon_\theta(z_t, t, \mathcal{E}(c_I), c_T) \right\|^2 \right]
\end{split}
  \label{eq:1}
\end{equation}
where $t \in T$ is timestep, $c_I$ is the image conditioning, $c_T$ is the text conditioning. During the testing phase, the prediction process of IP2P is as follows:
\begin{equation}
\begin{split}
    \tilde{\epsilon}_\theta(z_t, t, I, T)
    & = \epsilon_{\theta}(z_t, t, \emptyset_I,\emptyset_T) \\
    & + s_I(\epsilon_{\theta}(z_t, t, I, \emptyset_T)-\epsilon_{\theta}(z_t, t, \emptyset_I, \emptyset_T)) \\
    & + s_T (\epsilon_\theta(z_t, t, I, T) - \epsilon_\theta(z_t, t, I, \emptyset_T))
\end{split}
  \label{eq:2}
\end{equation}

\vspace{-15pt}
\subsection{Prompt Linear Interpolation}
Given two text prompts describing different colors (e.g.,``color the car light blue'' and ``color the car dark blue''), we first encode them using a pretrained text encoder CLIP~\cite{radford2021learning} to obtain prompt embeddings:
\begin{equation}
\begin{split}
    e_1 =  \text{CLIP}(\text{prompt}_1), e_2 = \text{CLIP}(\text{prompt}_2)
\end{split}
  \label{eq:2}
\end{equation}
By performing linear interpolation between $e_1$ and $e_2$, we obtain a sequence of intermediate embeddings representing a smooth semantic transition in the latent space:
\begin{equation}
\begin{split}
    e_i =  (1 - \alpha_i)e_1 + \alpha_ie_2 , \alpha_i \in [0, 1]
\end{split}
  \label{eq:2}
\end{equation}
where $i = \{1, 2, \dots, n\}$, $n$ is the number of sample. These intermediate embeddings are then used to guide the IP2P model, producing a series of images $\{I_i\}$ with gradually changing colors.

\vspace{-10pt}
\subsection{Color Mapper}
Although using interpolated embeddings as guidance yields images with varying colors, the RGB values in the generated images do not linearly correlate with the interpolation coefficient $\alpha$. This lack of a direct relationship makes it difficult to precisely control color through interpolation alone. To address this issue, we propose a learnable module, Color Mapper, which models the mapping between the image RGB values and the corresponding text embeddings.
For each interpolated embedding $e_i$ and its generated image $I_i$ , we sample a single pixel from the edited region (with predefined pixel coordinates), and record the RGB value $r_i$ at one pixel. The pair $(r_i, e_i)$, where $r_i \in \mathbb{R}^3$, forms one training sample.
The original embedding $e_i$ has a shape of $77 \times768$, we first apply PCA to compress it into a lower-dimensional vector $p_i \in \mathbb{R}^m$:
\begin{equation}
\begin{split}
     p_i = f_{PCA}(e_i)
\end{split}
  \label{eq:2}
\end{equation}
where $i = \{ 1, 2, \dots, n \}$, $f_{PCA}$ denotes the PCA dimensionality reduction function. Then, we design a multi-layer perceptron (MLP) to learn the mapping from RGB values to embedding space:
\begin{equation}
\begin{split}
     \hat{p}_i = f_{MLP}(r_i)
\end{split}
  \label{eq:2}
\end{equation}
The MLP is trained by minimizing the mean squared error (MSE) loss between the predicted and actual PCA-compressed embeddings:
\begin{equation}
\begin{split}
     L_{MSE} = \frac{1}{n}\sum_{i=1}^{n}\|\hat{p}_i - p_i \|_2^2
     \end{split}
  \label{eq:2}
\end{equation}
The predicted compressed embedding $\hat{e}_i $ is transformed back to the original embedding space using the inverse PCA $\hat{e}_i = f^{-1}_{PCA}(\hat{p}_i)$.
% \begin{equation}
% \begin{split}
%      \hat{e}_i = f^{-1}_{PCA}(\hat{p}_i)
% \end{split}
%   \label{eq:2}
% \end{equation}

\vspace{-10pt}
\subsection{Inference Stage}
During inference, the user provides a target RGB color and the original image. The Color Mapper module predicts the corresponding embedding $\hat{e}$, which is then fed into the IP2P model to generate an image that matches the desired color. To ensure that only the target object is edited without affecting unrelated regions of the image, we apply a binary mask $M$ at the first step of the diffusion process to restrict the editing area. The mask $M = \text{SAM}(I, T_{object})$ is obtained using the Segment-Anything Model~\cite{ravi2024sam} (SAM),
% \begin{equation}
% \begin{split}
%     M = \text{SAM}(I, T_{object})
% \end{split}
%   \label{eq:2}
% \end{equation}
where $T_{object}$ is the edited object. The final masked output is computed as:
\begin{equation}
\begin{split}
    \tilde{\epsilon}_\theta(z_t, t, I, T)
    & = \epsilon_{\theta}(z_t, t, \emptyset_I,\emptyset_T) \\
    & + s_I(\epsilon_{\theta}(z_t, t, I, \emptyset_T)-\epsilon_{\theta}(z_t, t, \emptyset_I, \emptyset_T))\\
    & + s_T (\epsilon_\theta(z_t, t, I, T) - \epsilon_\theta(z_t, t, I, \emptyset_T)) \odot M
\end{split}
  \label{eq:2}
\end{equation}

\vspace{-20pt}
\section{Experiments}
\label{sec:maintitle}

\begin{figure*}[t]
\begin{center}
  \includegraphics[width=0.85\linewidth]{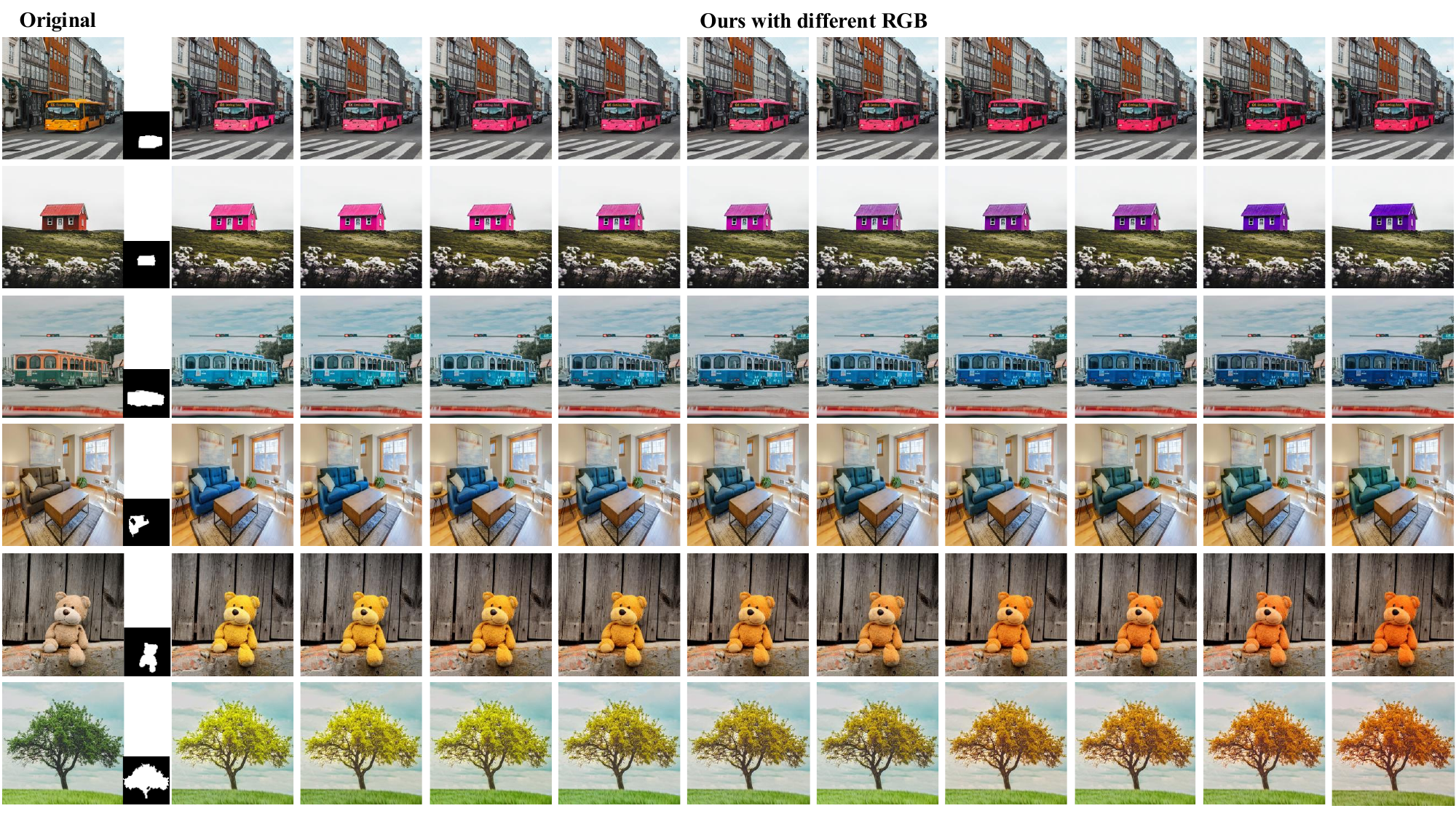}
\end{center}
\vspace{-15pt}
  \caption{Image color editing results with different RGB values. We selected several sets of linearly varying RGB for editing.}
  \label{fig3:main result}
  \vspace{-15pt}
\end{figure*}

\subsection{Implementation details}
We collected free-use images from Unsplash (\url{https://unsplash.com/}) and TEdBench~\cite{kawar2023imagic} as our test image. In our experiments, for the diffusion-based image editing component, we adopted the framework of InstructPix2Pix (IP2P) and used its pretrained weights. We employed the Euler ancestral sampler for a total of 100 denoising steps. The guidance scale parameters were set to $S_I = 1.5$ and $S_T = 7.5$. We used the Adam optimizer with a learning rate of 0.001 and trained Color Mapper for 500 epochs. For each image, we collected 30 embedding–RGB value pairs as training data. The PCA dimensionality reduction was set to 15 dimensions. All experiments were conducted on a single NVIDIA RTX 4090 GPU.

\vspace{-15pt}

\begin{figure}[t]
\centering %图片全局居中
\begin{minipage}[b]{0.35\linewidth} %所有minipage宽度之和要小于1，否则会自动变成竖排
    \centering %图片局部居中
        \includegraphics[width=0.95\textwidth]{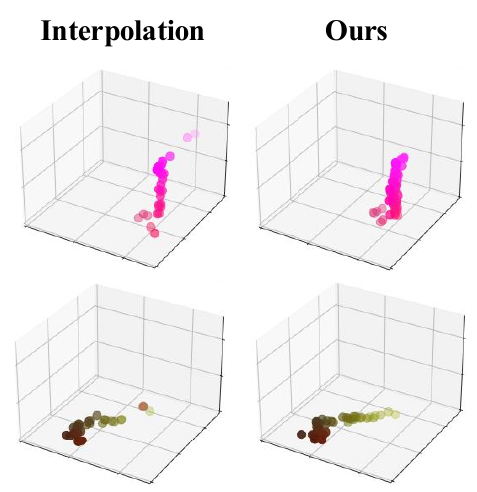} %此时的图片宽度比例是相对于这个minipage的，不是全局
    \caption{Edited image RGB distribution.}
    \label{fig4:RGB}
\end{minipage}
\begin{minipage}[b]{0.55\linewidth} %所有minipage宽度之和要小于1，否则会自动变成竖排
    \centering %图片局部居中
        \includegraphics[width=0.95\textwidth]{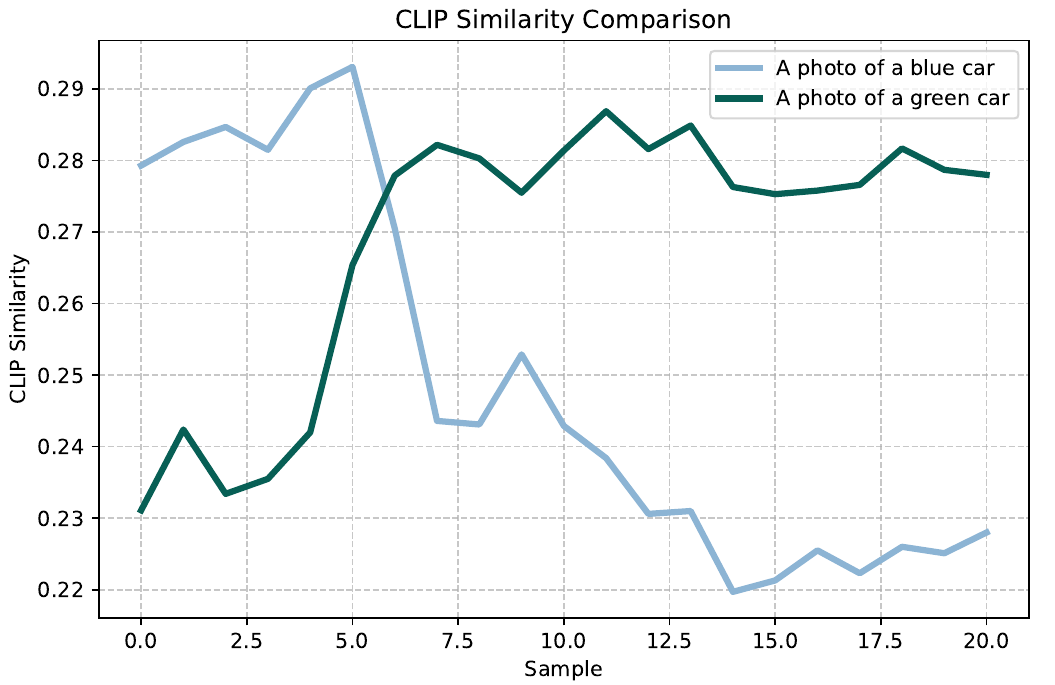}%此时的图片宽度比例是相对于这个minipage的，不是全局
    \caption{The CLIP similarity of a set of edited image.}
    \label{fig5:CLIP}
\end{minipage}
\vspace{-15pt}
\end{figure}

% {\bf Continuous Controllable Color Editing Results}
\subsection{Continuous Controllable Color Editing Results}
Based on the editing results of IP2P, we selected several RGB arrays as target color editing references and visualized the editing outcomes using the Color Mapper method, as shown in Fig.~\ref{fig3:main result}. The results demonstrate that:
i) Our framework can effectively isolate the target object for editing. Specifically, the SA segmentation model leverages object text prompts to automatically mask non-editing regions, preserving only the area of interest, which enhances the fidelity and consistency of the original image.
ii) By inputting RGB values, our method generates corresponding color embeddings to guide the image generation process, thereby enabling precise color control. Compared to conventional text-based prompts, RGB-based prompts offer a more intuitive and controllable editing interface.
iii) Our method supports progressive color editing by specifying continuously varying RGB arrays. As illustrated in Fig.~\ref{fig3:main result}, the color transitions smoothly from left to right without noticeable abrupt changes, demonstrating the effectiveness of our approach in achieving continuous color transformations.

% \noindent {\bf Quantitative comparisons}
\vspace{-10pt}
\subsection{Quantitative comparisons}
i) We extracted RGB color values from the edited regions of the images generated by both the interpolation method and our method (Ours). As shown in Fig.~\ref{fig4:RGB}, the RGB values in our generated images exhibit a more linear relationship. Compared to the interpolation method, our approach achieves smoother color transitions without abrupt changes or deviations in color.
ii) We constructed a set of RGB arrays representing a gradual transition from blue to green. For each generated image, we computed the CLIP similarity with two text prompts: ``a photo of a blue car'' and ``a photo of a green car''. As shown in Fig.~\ref{fig5:CLIP}, the similarity with the ``blue'' prompt decreases while the similarity with the ``green'' prompt increases along the sequence. This trend reflects the gradual color shift in the generated images and demonstrates the effectiveness of our method in controlling continuous color transformations.

\begin{figure}[t]
\begin{center}
  \includegraphics[width=0.9\linewidth]{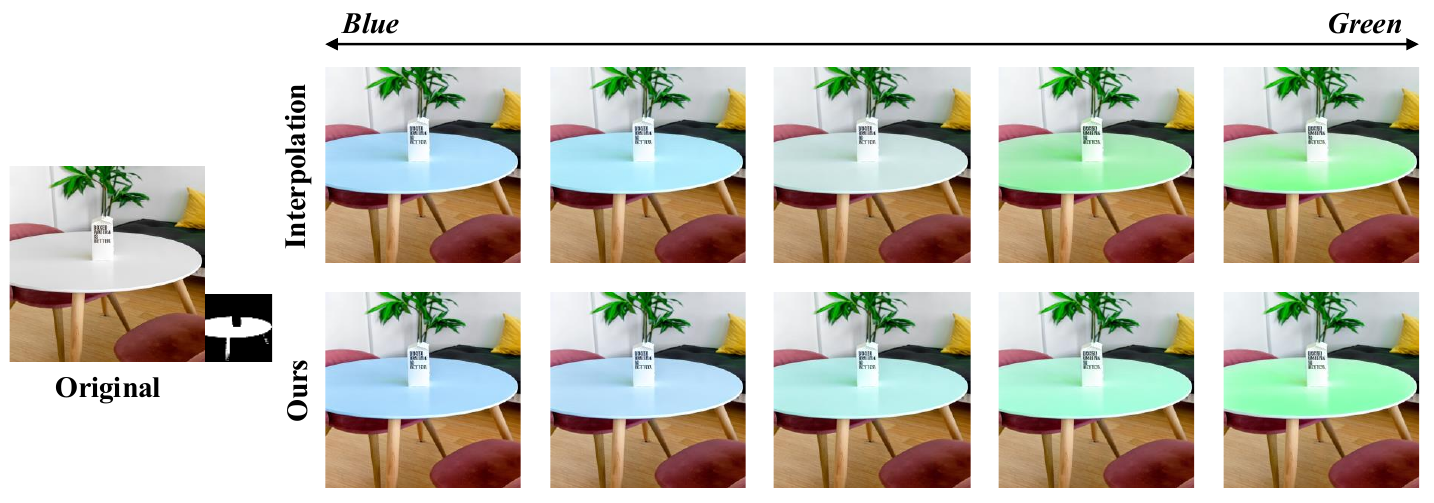}
\end{center}
\vspace{-15pt}
  \caption{Comparison with interpolation method.}
  \label{fig6:compare interp}
\vspace{-15pt}
\end{figure}

%{\bf User Study}
\begin{figure}[t]
\begin{center}
  \includegraphics[width=0.9\linewidth]{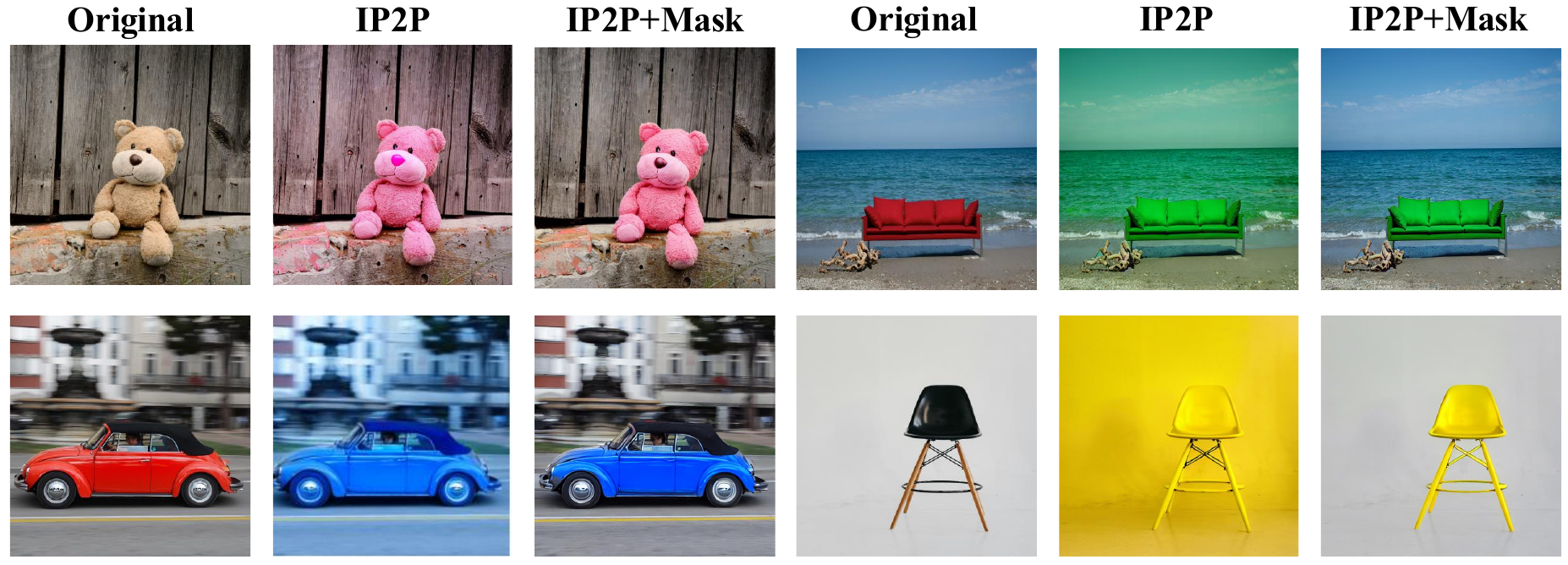}
\end{center}
\vspace{-15pt}
  \caption{Ablation study of mask generation.}
  \label{fig7:mask}
  \vspace{-15pt}
\end{figure}

\vspace{-15pt}
\subsection{Ablation Study}
% i) To demonstrate {\bf the effectiveness of our method in achieving smoother and more continuous color transitions}, we compared it with the interpolation-based method, as shown in Fig.~\ref{fig6:compare interp}. In the interpolation results, the color changes of the desk between the second and third images appear abrupt. Specifically, the brightness of the blue in the second image shifts too drastically, and the third image exhibits a grayish tone that deviates significantly from the target green color. In contrast, our method produces more gradual and natural color transitions across the sequence.
i) To demonstrate {\bf the effectiveness of our method in achieving smoother and more continuous color transitions}, we compared it with an interpolation-based method, as shown in Fig.~\ref{fig6:compare interp}. The interpolation results show abrupt color changes on the desk between the second and third images—specifically, an overly bright blue in the second image and a grayish tone in the third that deviates from the target green. In contrast, our method produces more gradual and natural transitions throughout the sequence.
% ii) {\bf To highlight the importance of the mask constraint}, we compared our method with the IP2P method without mask guidance. As illustrated in Fig.~\ref{fig7:mask}, IP2P tends to apply a color overlay across the entire image, affecting background regions unintentionally. Our method, on the other hand, preserves the background details effectively by applying color changes only to the intended object region, demonstrating better spatial precision in editing.
ii) {\bf To highlight the importance of the mask constraint}, we compared our method with IP2P without mask guidance. As shown in Fig.~\ref{fig7:mask}, IP2P often applies color overlays across the whole image, unintentionally altering the background. In contrast, our method restricts color changes to the target object, preserving background details and demonstrating superior spatial precision.

\vspace{-10pt}
\section{Conclusion}
\label{sec:typestyle}

This paper addresses the challenges of precise control and lack of continuity in color editing by proposing a color-controllable and continuous editing method based on color mapping. We introduce a learnable module, {\bf Color Mapper}, which learns the mapping between pixel-level RGB values and text embeddings, enabling users to perform continuous and controllable color edits directly through RGB values. Experimental results demonstrate that our method achieves region-controllable, color-controllable, and smoothly continuous color editing effects.
% This paper proposes a color-controllable and continuous editing method based on color mapping to address the challenges of precision and continuity in color editing. A learnable module, Color Mapper, maps pixel-level RGB values to text embeddings, enabling direct, continuous, and controllable color edits via RGB. Experiments show our method supports region- and color-controllable editing with smooth transitions.

\vspace{-15pt}
\section{Acknowledgements}
\label{}
This work was supported by the National Natural Science Foundation of China (Grant 62406171, 62225601, U23B2052), in part by the Beijing Natural Science Foundation Project No. L242025.

% \vfill\pagebreak


\bibliographystyle{IEEEbib}
\bibliography{strings,refs}

\end{document}